\documentclass[letterpaper]{article} 
\usepackage{aaai2026}  
\usepackage{times}  
\usepackage{helvet}  
\usepackage{courier}  
\usepackage[hyphens]{url}  
\usepackage{graphicx} 
\usepackage{natbib}  
\usepackage{caption} 
\usepackage{algorithm}
\usepackage{algorithmic}
\usepackage{algorithmic}
\usepackage{multirow} 
\usepackage{enumerate}
\usepackage{amsfonts}
\usepackage{amsmath}
\usepackage{mathrsfs}
\usepackage{booktabs}
\usepackage{array} 
\usepackage[table]{xcolor}
\usepackage{newfloat}
\usepackage{listings}
\DeclareCaptionStyle{ruled}{labelfont=normalfont,labelsep=colon,strut=off} 
\floatstyle{ruled}
\newfloat{listing}{tb}{lst}{}
\floatname{listing}{Listing}
\title{CD-DPE: Dual-Prompt Expert Network Based on Convolutional Dictionary Feature Decoupling for Multi-Contrast MRI Super-Resolution}
\author{
    Xianming Gu\textsuperscript{\rm 1},
    Lihui Wang\textsuperscript{\rm 1}\thanks{Corresponding author.},
    Ying Cao\textsuperscript{\rm 1},
    Zeyu Deng\textsuperscript{\rm 1},
    Yingfeng Ou\textsuperscript{\rm 1},
    Guodong Hu\textsuperscript{\rm 1},
    Yi Chen\textsuperscript{\rm 1,2}
}
\affiliations{
    \textsuperscript{\rm 1}Key Laboratory of Advanced Medical Imaging and Intelligent Computing of Guizhou Province, \\Engineering Research Center of Text Computing \& Cognitive Intelligence, Ministry of Education,\\ College of Computer Science and Technology, Guizhou University, Guiyang, China\\

	\textsuperscript{\rm 2}The D-Lab, Department of Precision Medicine, GROW-School for Oncology and Reproduction,\\ Maastricht University, 6200 MD Maastricht, the Netherlands\\
	xianming\_gu@foxmail.com, lhwang2@gzu.edu.cn
}

\usepackage{bibentry}

\begin{document}

\maketitle

\begin{abstract}
Multi-contrast magnetic resonance imaging (MRI) super-resolution intends to reconstruct high-resolution (HR) images from low-resolution (LR) scans by leveraging structural information present in HR reference images acquired with different contrasts. This technique enhances anatomical detail and soft tissue differentiation, which is vital for early diagnosis and clinical decision-making. However, inherent contrasts disparities between modalities pose fundamental challenges in effectively utilizing reference image textures to guide target image reconstruction, often resulting in suboptimal feature integration. To address this issue, we propose a dual-prompt expert network based on a convolutional dictionary feature decoupling (CD-DPE) strategy for multi-contrast MRI super-resolution. Specifically, we introduce an iterative convolutional dictionary feature decoupling module (CD-FDM) to separate features into cross-contrast and intra-contrast components, thereby reducing redundancy and interference. To fully integrate these features, a novel dual-prompt feature fusion expert module (DP-FFEM) is proposed. This module uses a frequency prompt to guide the selection of relevant reference features for incorporation into the target image, while an adaptive routing prompt determines the optimal method for fusing reference and target features to enhance reconstruction quality. Extensive experiments on public multi-contrast MRI datasets demonstrate that CD-DPE outperforms state-of-the-art methods in reconstructing fine details. Additionally, experiments on unseen datasets demonstrated that CD-DPE exhibits strong generalization capabilities.
\end{abstract}

\begin{links}
    \link{Code and Supplementary Materials}{https://github.com/xianming-gu/CD-DPE}
\end{links}

\section{Introduction}

\begin{figure}[!t]	\centering{\includegraphics[width=\linewidth]{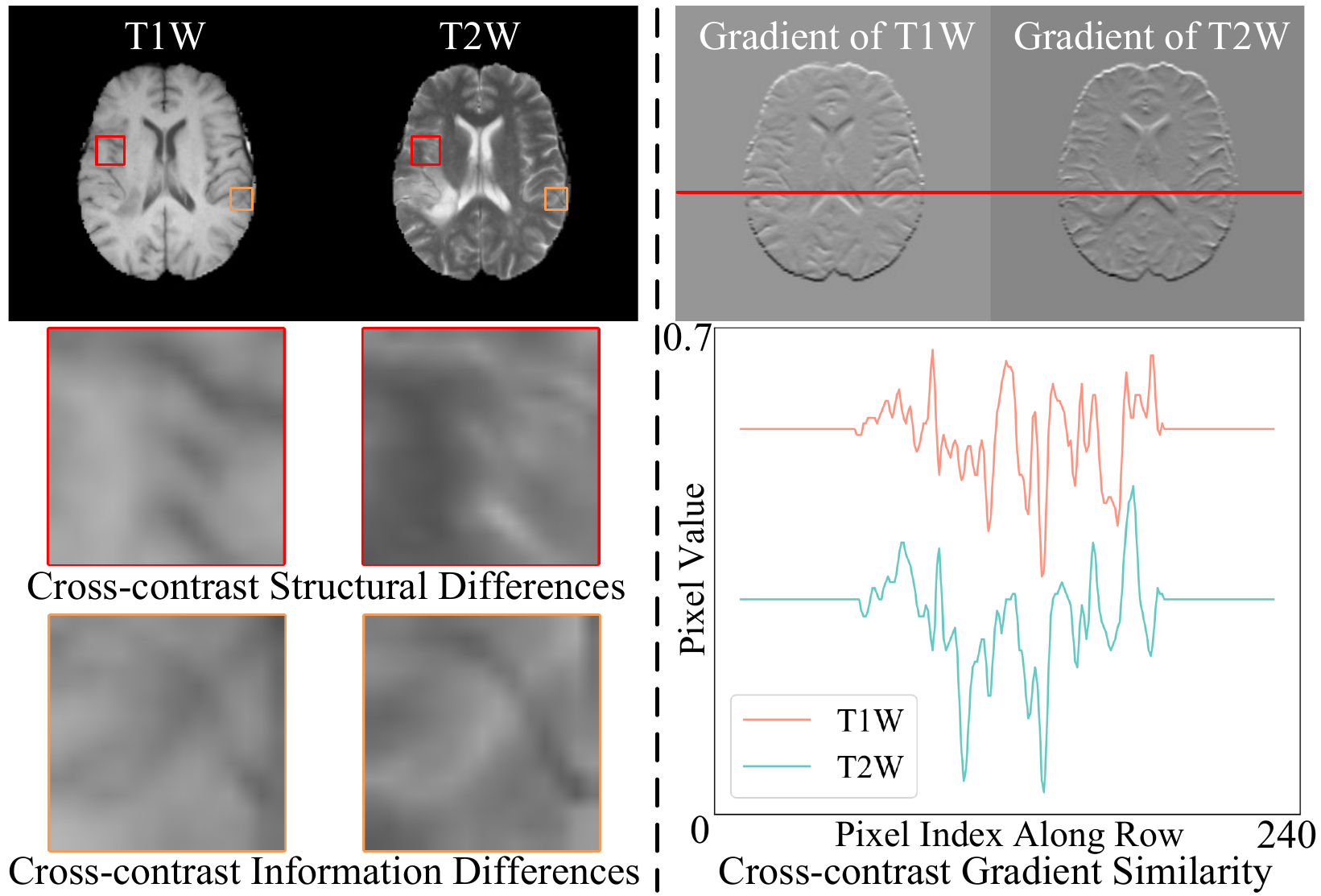}}
	\caption{Structural disparities and shared information across multi-contrast MRI.}
	\label{fig0}
\end{figure}

Magnetic resonance imaging (MRI) provides substantial clinical benefits as a non-invasive modality that avoids ionizing radiation exposure \cite{de2016accuracy,umirzakova2024medical,zhao2024mri,muhammad2020deep}. However, obtaining high-resolution (HR) MRI images faces inherent limitations due to physical imaging constraints and physiological factors \cite{feng2021task, lyu2020multi, feng2022multimodal, vakli2023automatic}. Super-resolution (SR) techniques overcome this challenge by reconstructing HR images from low-resolution (LR) acquisitions \cite{zhao2019channel,huang2024mftn}, thereby improving diagnostic accuracy. In clinical practice, MRI protocols typically acquire multiple contrast-weighted sequences (e.g., T1-weighted (T1W), T2-weighted (T2W), and proton density-weighted (PD)) to generate complementary diagnostic images. This presents an opportunity where rapidly acquired HR references (e.g., T1W) could potentially enhance LR targets requiring longer scan times (e.g., T2W). However, even with aligned multi-contrast images, structural and informational disparities persist due to contrast variations (Figure \ref{fig0}). Consequently, effectively utilizing the shared information from contrast-mismatched HR references remains a significant challenge in current multi-contrast MRI SR approaches \cite{zhao2019applications,granziera2015multi}.

Early CNN-based approaches \cite{lyu2020multi,feng2021multi,liu2023flexible,feng2024exploring} employ simple fusion strategies, either concatenating reference and target images as model inputs or integrating their high-level features. Although computationally efficient, such direct concatenation fails to capture complex cross-contrast dependencies, limiting its effectiveness in modeling structural relationships between reference and target domains and therefore resulting in unsatisfactory reconstruction results with blurred details. 
Transformer-based SR methods have attracted considerable research interest due to their capabilities in long-range dependency modeling and flexible feature integration. These approaches utilize diverse attention mechanisms \cite{feng2022multimodal, li2022wavtrans, huang2023accurate} to fuse features from reference and target images. Despite achieving notable performance, they suffer from two critical limitations: inherent constraints in reconstructing high-frequency details from very low-resolution inputs degrade output fidelity, while intensive computational demands result in high memory consumption and prolonged processing time \cite{vaswani2017attention,liu2021swin}.
To mitigate the aforementioned fusion limitations, several studies employ handcrafted strategies, such as multi-scale context aggregation \cite{li2022transformer}, texture search \cite{ruan2024reference}, and neighborhood-guided aggregation \cite{chen2025multi}, which effectively enhance texture details in the target image. Nevertheless, these manually designed approaches inherently exhibit constrained generalization capability and adaptability \cite{yang2022model,yang2023mgdun}.

Recent studies decompose HR reference images into distinct components to guide LR target reconstruction. Specifically, Li et al. \cite{li2024rethinking} separate HR references into high-frequency priors and structural features, subsequently fusing them with LR features via a diffusion model to achieve distortion-free reconstruction. Although inference efficiency improved compared to standard diffusion models, it remains suboptimal \cite{mao2023disc}. Instead, Lei et al. \cite{lei2023decomposition,lei2025robust} decompose images into common and unique components, transferring exclusively common features from HR references to LR reconstruction targets to minimize redundancy interference. This kind of methods provide an effective means for multi-contrast SR reconstruction, however, they lack rigorous constraints on the decomposition and fusion mechanisms between common and target-unique features, risking significant degradation if common features become over-smoothed or fusion strategies are inappropriate.

To address these challenges, we propose a Dual-Prompt Expert network based on a Convolutional Dictionary feature decoupling strategy (CD-DPE) for multi-contrast MRI SR reconstruction. It first extract the unique and common features of HR reference and LR target images, and then using an expert model to fuse and reconstruct the HR target image. The main contributions of this work are summarized as follows:
\begin{enumerate}[1)]
	\item We propose a convolutional dictionary feature decoupling module (CD-FDM) to effectively separate multi-contrast MR images into distinct cross-contrast unique features and intra-contrast common features, eliminating redundant information interference while preserving essential structural details for improved super-resolution reconstruction.	
	\item A novel Dual-Prompt Feature Fusion Expert Module (DP-FFEM) is introduced, which leverages frequency-aware and routing-adaptive prompts to intelligently fuse HR-LR features, dynamically optimizing both feature selection and fusion rules for enhanced reconstruction.
	\item Extensive experiments on two public multi-contrast MRI datasets demonstrate that our method achieves state-of-the-art performance compared to existing approaches. Additionally, CD-DPE demonstrated strong generalization capabilities when validated on unseen datasets.
\end{enumerate}

\begin{figure*}[!t]
	\centerline{\includegraphics[width=0.95\linewidth]{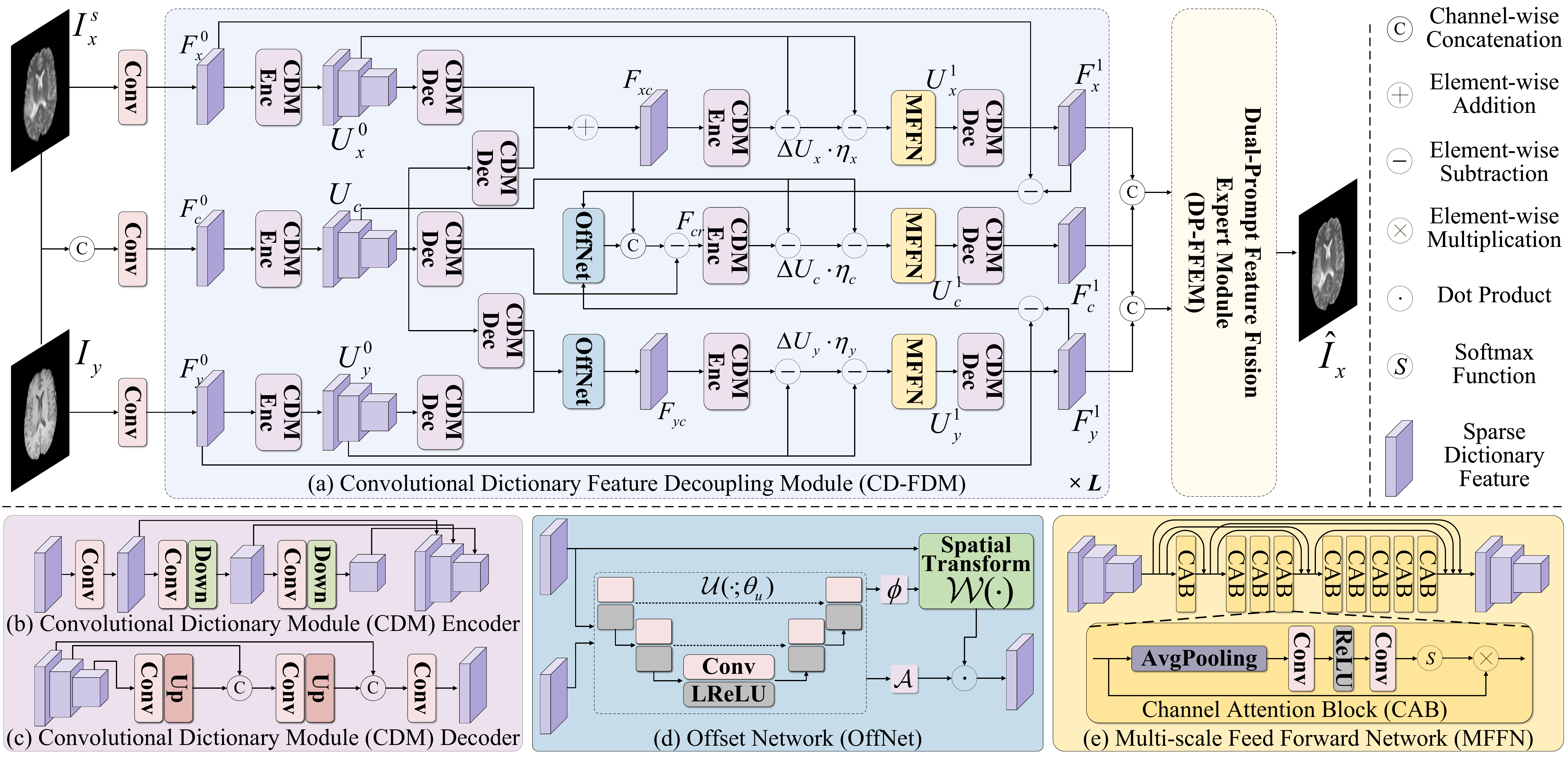}}
	\caption{The architecture of dual-prompt expert network based on convolutional dictionary feature decoupling (CD-DPE).}
	\label{fig_Framework}
\end{figure*}

\section{Methods}
\label{sec_method}
\subsection{Problem Formulation}
\label{sec_Problem}
The goal of multi-contrast MRI super-resolution is to reconstruct a high-resolution, fully sampled target image \(\hat{I}_x \in \mathbb{R}^{H \times W}\) from its LR undersampled counterpart \(I_x \in \mathbb{R}^{H/s \times W/s}\), guided by an HR cross-contrast reference image \(I_y \in \mathbb{R}^{H \times W}\). Formally, the reconstruction can be expressed as:  
\begin{equation}
\hat{I_x} = f(I^s_x|I_y;\theta_f),
\label{eq1}
\end{equation}
where \(I^s_x \in \mathbb{R}^{H \times W}\) denotes the upsampled LR image with an upscaling factor \(s\), and \(f(\cdot; \theta_f)\) represents the reconstruction function parameterized by learnable weights \(\theta_f\).  

Effectively extracting and leveraging information from the reference image $I_y$ to guide the reconstruction of $\hat{I}_x$ is therefore critical.
Since both images originate from the same anatomical structure, they inherently share common features. However, differences in acquisition protocols introduce modality-specific characteristics, leading to distinct contrasts. To model this, multi-contrast MRI images can be decomposed into unique and shared components:  
\begin{equation}
	\begin{aligned}
		I^s_x&=\sum_{j}^{J}u^x_j\otimes \theta^x_d + c_j\otimes \theta^c_d,\\
		 I_y&=\sum_{j}^{J}u^y_j\otimes \theta^y_d + c_j\otimes \theta^c_d,
	\end{aligned}
	\label{eq2}
\end{equation}
where $u^x_j$ and $u^y_j$ denote the unique sparse representations of the LR target image and HR reference image, respectively, while $c_j$ represents the common sparse representations. The index $j=\{1,2,3,...,J\}$  corresponds to the feature scales, and $\otimes$ denotes convolution with dictionary filters $\{ \theta^x_d,\theta^y_d,\theta^c_d \}$. How to effectively extract the unique and common features from the input images  is challenging. Assuming that the dictionary filters are known, then the unique and common sparse representations can be optimized by:
\begin{equation}
	\begin{aligned}
		\min_{\{u^x_j,u^y_j,c_j\}} &\frac{1}{2}\lVert{I^{hr}_x}- \sum_{j}^{J}u^x_j\otimes \theta^x_d + c_j\otimes \theta^c_d\rVert^2_F\\
		+&\frac{1}{2}\lVert{I_y}- \sum_{j}^{J}u^y_j\otimes \theta^y_d + c_j\otimes \theta^c_d\rVert^2_F\\
		+&\sum_{j}^{J}\varphi(u^x_j,c_j)+\varphi(u^y_j,c_j),
	\end{aligned}
	\label{eq3}
\end{equation}
where $I^{hr}_x$ is the HR targt image and $\varphi(\cdot)$ represents the optimization function for learning unique and common features. 

By stacking the multi-scale features  $\{u^x_j,u^y_j,c_j\}$ across all scales $j=1,2,...,J$, we obtain the sparse representations $\{U_x, U_y, U_c\}$, where $U_x$ and  $U_y$ encode the modality-unique features of the target and reference images, respectively, $U_c$ captures the shared anatomical structure across contrasts.
These representations are then integrated through a fusion and reconstruction model $g(\cdot,\theta_r)$ parameterized by weights $\theta_r$, to generate the SR output of the LR target image:
\begin{equation}
\hat{I}_x= g(U_x, U_y, U_c; \theta_r)\label{eq4}
\end{equation}
The reconstruction network can then be optimized by minimizing the following objective:
\begin{equation}
		\min_{\theta_r} \frac{1}{2}\lVert g(U_x, U_y, U_c; \theta_r)- {I^{hr}_x}\rVert^2_F.
		\label{eq5}
\end{equation}
 
Through joint optimization of Eq. (\ref{eq3}) and Eq.  (\ref{eq5}), the model effectively learns the optimal solution for multi-contrast MRI super-resolution reconstruction, where the critical challenges involve effectively extracting both unique and common features, as well as fusing them for image reconstruction.

\subsection{Network Architecture}

To address these two challenges, we propose a CD-DPE network (Figure \ref{fig_Framework}), comprising two core modules: a CD-FDM for extracting common and unique features, and a DP-FFEM that adaptively fuses features of LR target and HR reference images for reconstruction. The specific details of each module will be described in the following sections.

\subsubsection{Structure of CD-FDM}
CD-FDM uses convolutional dictionaries \cite{gregor2010learning} to capture multi-scale unique and common feature representations, as shown in Figure \ref{fig_Framework}(a). First, the upsampled LR target image $I_x^s$, HR reference image $I_y$, and their concatenation are separately fed into convolutional layers to extract initial unique and common feature representations $\{F_x^0, F_y^0, F_c^0\}$. These features are then passed through a convolutional dictionary module encoder (CDM$_E$, Figure \ref{fig_Framework}(b)) to obtain multi-scale sparse representations $\{U_x^0, U_y^0, U_c^0\}$. According to the idea of unfold learning, these unique and common multi-scale sparse representations are updated with an iterative method, formulated as: 
\begin{equation}
\small
\begin{aligned}
	F_{xc}&=\text{CDM}_D(U_x^{l-1})+\text{CDM}_D(U_c^{l-1}),\\
	\Delta U_x &= U_x^{l-1} - \text{CDM}_E(F_{xc}),\\
	U^l_x & = \text{Prox}(U_x^{l-1} -\eta_x \Delta U_x), l=1,2,...,L
\end{aligned}
	\label{eq6}
\end{equation}
where $\text{CDM}_D$ indicates the inverse dictionary operation, implemented with a decoder structure (Figure \ref{fig_Framework}(c)). The proximal operation ($\text{Prox}$) is realized with multi-scale feed forward network (MFFN, Figure \ref{fig_Framework}(e)).
The unique features of input LR image $I_x^s$ is then obtained by: 
\begin{equation}
		F_x^l= \text{CDM}_D(U^l_x).
	\label{eq7}
\end{equation}

Similarly, the unique features of HR reference image $I_y$ are extracted following a comparable process. However, to address potential misalignments between the reference and target images, an offset network (OffNet) implemented with spatial transformation \cite{chen2022transmorph,huang2021coarse} is introduced, as illustrated in Figure \ref{fig_Framework}(d). OffNet employs a lightweight U-Net \cite{ronneberger2015u} $\mathcal{U}(\cdot;\theta_u)$ to learn both displacement field $\phi$ and corresponding feature representations $\mathcal{A}$. A spatial transformation module $\mathcal{W}(\cdot)$ is then applied to align the reference features with the target image, which can be expressed as:
\begin{equation}
\small
\begin{aligned}
       \phi,\mathcal{A} &= \mathcal{U}([\text{CDM}_D(U_c^{l-1}),\text{CDM}_D(U_y^{l-1})];\theta_u),\\
      F_{yc}&= \mathcal{W}(\text{CDM}_D(U_c^{l-1}), \phi)\odot\mathcal{A},\\       
      \Delta U_y &= U^{l-1}_y-\text{CDM}_E\left(\text{CDM}_D(U_y^{l-1})+F_{yc}\right),\\
      U^l_y & = \text{Prox}(U_y^{l-1} -\eta_y \Delta U_y), l=1,2,...,L
	\label{eq8}
\end{aligned}
\end{equation}
where $[\cdot,\cdot]$ represents the concatenation operation along the channel dimension. From the unique sparse representations $U_y^l$, the unique features of HR reference image can also be obtained by: 
\begin{equation}
		F_y^l= \text{CDM}_D(U^l_y).
	\label{eq9}
\end{equation}

To update the common features between the target and reference images, we first refine the common features by subtracting the residual features derived from reference and target unique features. Note that, to avoid any misalignment between reference and target image, the residuals of target unique features are first warped through the OffNet, the refined common features $F_{cr}$ can be written as: 
\begin{equation}
  \small
  \begin{aligned}
	\phi^\prime,\mathcal{A}^\prime &= \mathcal{U}([F_y^l-F_y^{l-1},F_x^l-F_x^{l-1}];\theta_u),\\		
	F_{cr}&=\text{CDM}_D(U_c^{l-1})-[\mathcal{W}(F_y^l-F_y^{l-1}, \phi^\prime)\odot\mathcal{A}^\prime,F_x^l-F_x^{l-1}]\\
   \label{eq10}
   \end{aligned}
\end{equation}
From the refined common features, the common feature sparse representations can be formulated as:
\begin{equation}
\small
\begin{aligned}
	\Delta U_c &= U_c^{l-1} - \text{CDM}_E(F_{cr}),\\
	U_c^{l} &=\text{Prox}(U_c^{l-1}-\eta_c \Delta U_c), l=1,2,...,L
	\label{eq11}
\end{aligned}
\end{equation}
The common features can be accordingly updated with:
\begin{equation}
		F_c^l= \text{CDM}_D(U^l_c).
	\label{eq12}
\end{equation}

In this work, CD-FDM is repeated $L$ times to optimize the unique and common features, the final unique and common features of LR target and HR reference images can be noted as $F_x^L$, $F_y^L$, and $F_c^L$, respectively.

\subsubsection{Structure of DP-FFEM}

\begin{figure}[!t]
	\centerline{\includegraphics[width=\linewidth]{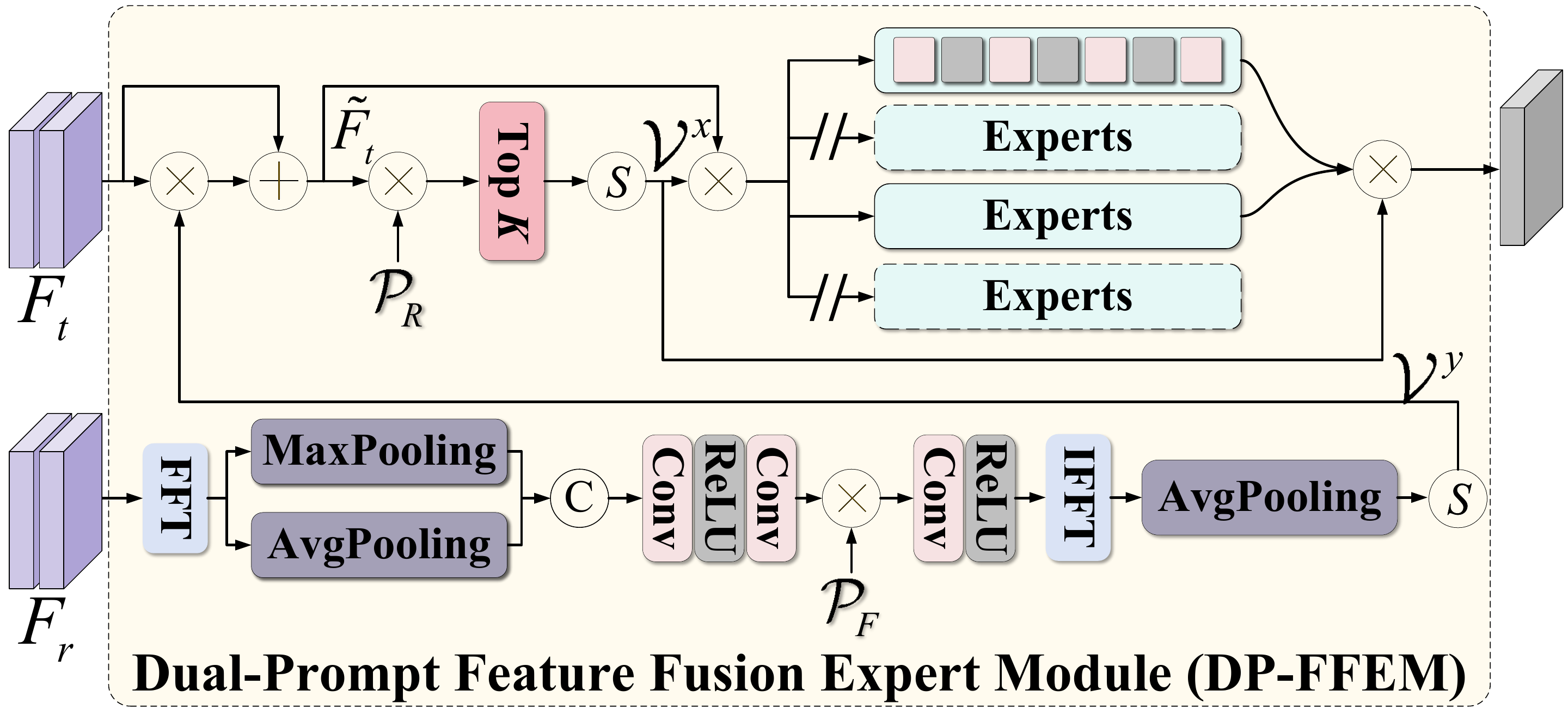}}
	\caption{The architecture of dual-prompt feature fusion expert module (DP-FFEM).}
	\label{fig_Fusion}
\end{figure}
To achieve effective cross-modality guidance and improve super-resolution performance via reference image integration, we propose DP-FFEM. As illustrated in Figure \ref{fig_Fusion}, it leverages a novel dual-prompt mechanism to comprehensively guide the target image reconstruction process through multi-level feature interaction.
This module first establishes dual feature representations: the reference representation integrates modality-specific and shared features via $F_r = [ F_y^L, F_c^L]$, while the target representation combines its own features with shared patterns through $F_t = [ F_x^L, F_c^L]$. 
Despite limited statistical correlation between shared features $F_c^L$ and reference-specific features $F_y^L$, their spatial attention map provides crucial reconstruction guidance. Specifically, such attention map identifies semantically consistent regions where modality-specific features should adaptively align with shared representations, thereby enhancing feature compatibility throughout the reconstruction pipeline.
 In our implementation, this attention $\mathcal{V}^y$ is  dynamically regulated through a learnable frequency prompt designed to capture and emphasize crucial structural patterns in the feature space, that means:

\begin{equation}
	\mathcal{V}^y = f_{\phi_1}(\mathscr{F}(F_r), \mathcal{P}_F)
	\label{eq:22}
\end{equation}
where $\mathscr{F}(\cdot)$ denotes the Fourier transform, $\mathcal{P}_F$ a learnable frequency prototype, and the detailed structure of $f_{\phi_1}$ can be found in Figure \ref{fig_Fusion}. Given that the target and reference images capture the same underlying scene, we transfer the attention maps $\mathcal{V}^y$ derived from the reference image's unique and common features to guide the feature enhancement of the target representation $F_t$,
\begin{equation}
		\tilde{F}_t = F_t\otimes \mathcal{V}^y+ F_t.
	\label{eq:26}
\end{equation}
 Through this attention-aware feature enhancement, we ensure that the target reconstruction preserves spatial coherence while effectively incorporating complementary information from the reference image. Subsequently, a learnable adaptive routing prompt $\mathcal{P}_R \in \mathbb{R}^{(C\times H\times W)\times E}$ is introduced to guide dynamic routing within the expert network for fusion. Specifically, $\mathcal{P}_R$ is multiplied with the target features to generate routing logits, from which the Top-$K$ operator \cite{shazeer2017outrageously,cao2023multi} selects the most relevant $K$ expert branches. These selections are then normalized using the Softmax function to produce routing weights $\mathcal{V}^x$, formulated as:
\begin{equation}
		\mathcal{V}^x=\text{Softmax}(\text{Top}K(\text{Flatten}(\tilde{F}_t)\otimes \mathcal{P}_R)).
		\label{eq:28}
\end{equation}
The final reconstruction result is a linearly weighted combination of the $K$ most relevant outputs from the $E$ experts $\mathcal{E}(\cdot)$ and the corresponding routing weights, formulated as:
\begin{equation}
		\hat{I}_x=\sum_{i=1}^{E}\mathcal{V}^x\cdot \mathcal{E}_i(\tilde{F}_t\cdot \mathcal{V}^x)
		\label{eq:29}
\end{equation}

\begin{table*}[!t]
	\centering
\begin{tabular}{lccccccc}
\toprule
\multirow{2}[4]{*}{Methods} & \multicolumn{2}{c}{BraTS2018 2$\times$} & \multicolumn{2}{c}{BraTS2018 4$\times$} & \multicolumn{3}{c}{Model Efficiency} \\
\cmidrule{2-8}      & PSNR$\uparrow$  & SSIM$\uparrow$  & PSNR$\uparrow$  & SSIM$\uparrow$  & Params(M) & FLOPs(G) & Times(s) \\
\midrule
WavTrans & 39.7915$\pm$2.66 & 0.9874$\pm$0.01 & 34.8263$\pm$2.53 & 0.9677$\pm$0.01 & 10.015  & 216.150  & 0.203 \\
SANet & 36.2761$\pm$2.35 & 0.9839$\pm$0.01 & 32.0269$\pm$2.18 & 0.9569$\pm$0.01 & 11.857  & 259.573  & 0.041 \\
DiffMSR & \multicolumn{1}{c}{/}      & \multicolumn{1}{c}{/}      & 31.3899$\pm$2.55 & 0.9638$\pm$0.01 & 6.603  & 302.008  & 0.358 \\
DANCE & 32.5425$\pm$2.57 & 0.9804$\pm$0.01 & 31.7239$\pm$2.47 & 0.9645$\pm$0.01 & 43.273  & 57.504  & 0.089 \\
A2-CDic & \underline{40.4682$\pm$2.68} & \underline{0.9883$\pm$0.01} & \underline{35.6983$\pm$2.60} & \underline{0.9704$\pm$0.01} & 10.066  & 831.073  & 0.114 \\
\rowcolor[rgb]{ .906,  .902,  .902} CD-DPE & \textbf{40.7047$\pm$2.49} & \textbf{0.9885$\pm$0.01} & \textbf{36.0017$\pm$2.33} & \textbf{0.9716$\pm$0.01} & 11.705  & 426.099  & 0.061 \\
\bottomrule
\end{tabular}%
	\caption{Quantitative comparison results and model efficiency of multi-contrast MRI super-resolution on BraTS2018 dataset. \textbf{Bold} indicates the optimal value, while \underline{underline} indicates the second-best value.}
	\label{tab_Results_BraTS}%
\end{table*}%

\subsection{Loss Function}
The total loss function consists of three parts, including consistency loss, decoupling loss and reconstruction loss. The consistency loss constrains the $L_1$ distance between the image and the features derived from the unique and common feature combinations, formulated as:
\begin{equation}
		\mathcal{L}_{fc} = \lVert{I^{hr}_x}- (F_x^L+F_c^L)\rVert_1+\lambda_{y}\lVert{I_y}- (F_c^L+F_y^L)\rVert_1,
		\label{eq:30}
\end{equation}
where $\lambda_{y}=0.01$ is a weighting factor that balances the contributions of the two terms.

For the decoupling loss, it requires less dependence between decoupled unique and common features, that means, 
\begin{equation}
		\mathcal{L}_{mi} = \text{MI}(F_c^L, F_x^L)+\text{MI}(F_c^L,F_y^L).
		\label{eq:31}
\end{equation}
where $\text{MI}$ indicates the mutual information. 
The reconstruction loss is used to supervise the content consistency of the reconstructed image, formulated as:
\begin{equation}
		\mathcal{L}_{rec} = \lVert\hat{I_x}- I^{hr}_x\rVert_1.
		\label{eq:32}
\end{equation}

Finally, the overall loss function is formulated as a weighted combination of the aforementioned components. The network is trained and optimised in an end-to-end manner, and the total loss is defined as:
\begin{equation}
		\mathcal{L} = \mathcal{L}_{rec}+\lambda_1\mathcal{L}_{fc}+\lambda_2\mathcal{L}_{mi},
		\label{eq:33}
\end{equation}
where $\lambda_1$ and $\lambda_2$ are trade-off parameters that balance the relative contributions of the three loss components.

\begin{figure*}[!h]
	\centerline{\includegraphics[width=0.82\linewidth]{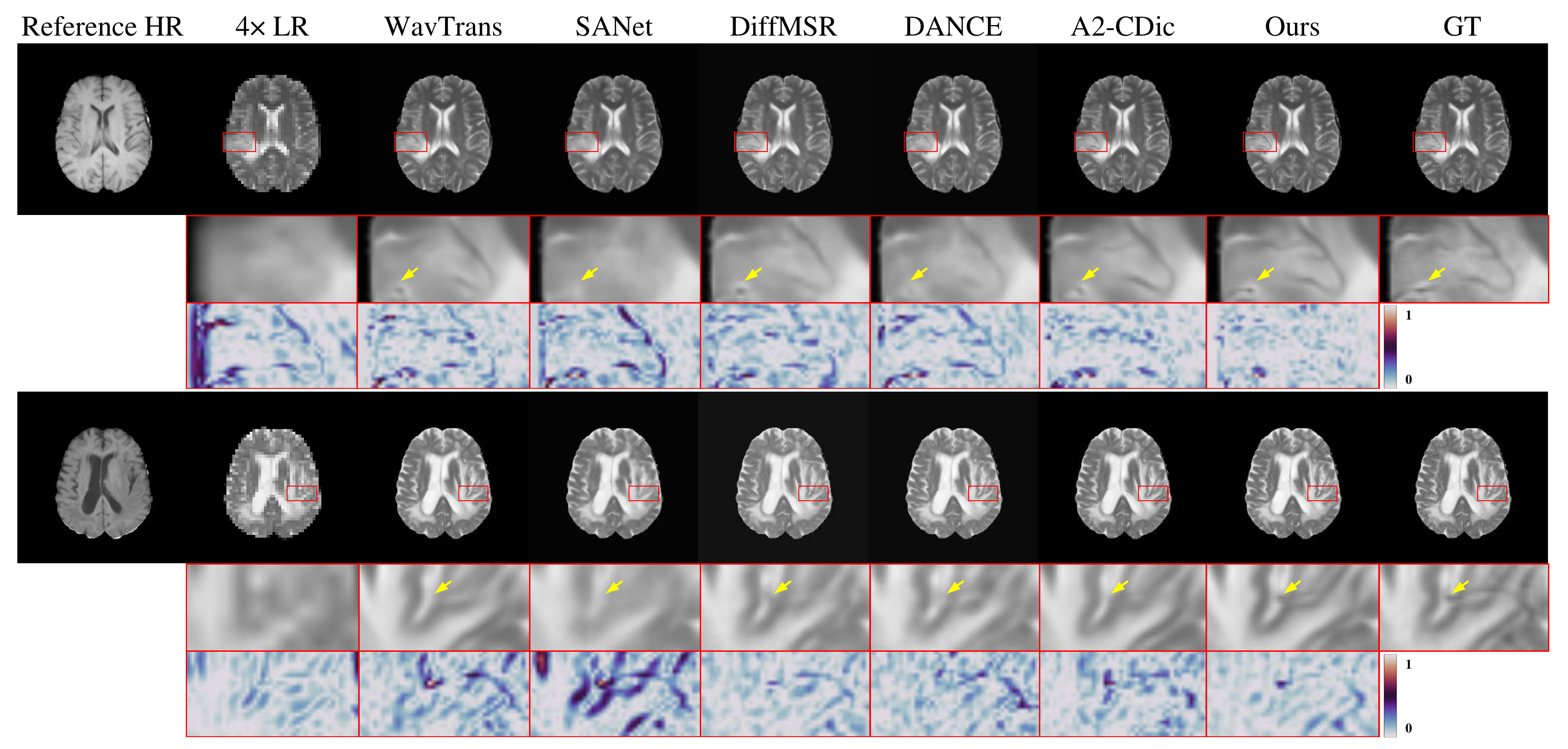}}
	\caption{Qualitative comparison of various methods on the BraTS2018 dataset with $4\times$ SR. The yellow arrows indicate areas with significant differences, which are enlarged and shown with residual plots compared to ground-truth (GT).}
	\label{fig_Results_BraTS}
\end{figure*}

\section{Experiments}
\label{sec_Experiments}
\subsection{Datasets}
We evaluate our method on two public datasets, including BraTS2018 \cite{menze2014multimodal} and IXI (available at https://brain-development.org/ixi-dataset/). BraTS2018 contains 285 preprocessed, spatially aligned multi-contrast MRI scans. We use the central 50 slices (240$\times$240) from each scan, with T1W as the reference to reconstruct T2W, yielding over 11,000 training and 2,800 test pairs. The IXI dataset includes 576 similarly preprocessed scans, from which 50 central slices (256$\times$256) are selected. PD is used to guide T2W reconstruction, resulting in over 23,000 training and 5,700 test pairs. LR images are generated by downsampling HR images by a factor of $2\times$ or $4\times$, and are upsampled via interpolation for network input.

\subsection{Implement Details and Metrics}
We implemented these models using the PyTorch framework and trained them on NVIDIA RTX A6000 GPU with single-card 48GB memory for 50 epochs. The batch size was set to 4. We used the Adam optimiser \cite{kingma2014adam} with a learning rate of $1\times10^{-4}$. In CD-DPE, the number of convolutional kernels in the initial convolutional layer is 64. In CD-FDM, the number of iterations $L$ is set to 3, the levels of the CDMs are set to 3, and the number of channels is 64, 96, and 128 from the first to the third level, respectively. The initial values of modulation parameters $\eta_x$, $\eta_y$, and $\eta_c$ are set to 0.01 and are updated during the learning process. In DP-FFEM, the number of experts $E$ in the expert network is set to 4, and $K$ is set to 2. In the loss function, $\lambda_1$ and $\lambda_2$ are set to 1 and 0.1, respectively.

We conducted comparative experiments with the previous five methods, including WavTrans \cite{li2022wavtrans}, SANet \cite{feng2024exploring}, DiffMSR \cite{li2024rethinking} (only for $4\times$ SR), DANCE \cite{chen2025multi} and A2-CDic \cite{lei2025robust}. All of them are evaluated on the BraTS2018 and IXI datasets using $2\times$ and $4\times$ super-resolution magnification factors. Model performance was quantitatively assessed using peak signal-to-noise ratio (PSNR) and structural similarity index measure (SSIM). Higher PSNR and SSIM values indicate better super-resolution effects.

\begin{figure*}[!h]	\centering{\includegraphics[width=0.82\textwidth]{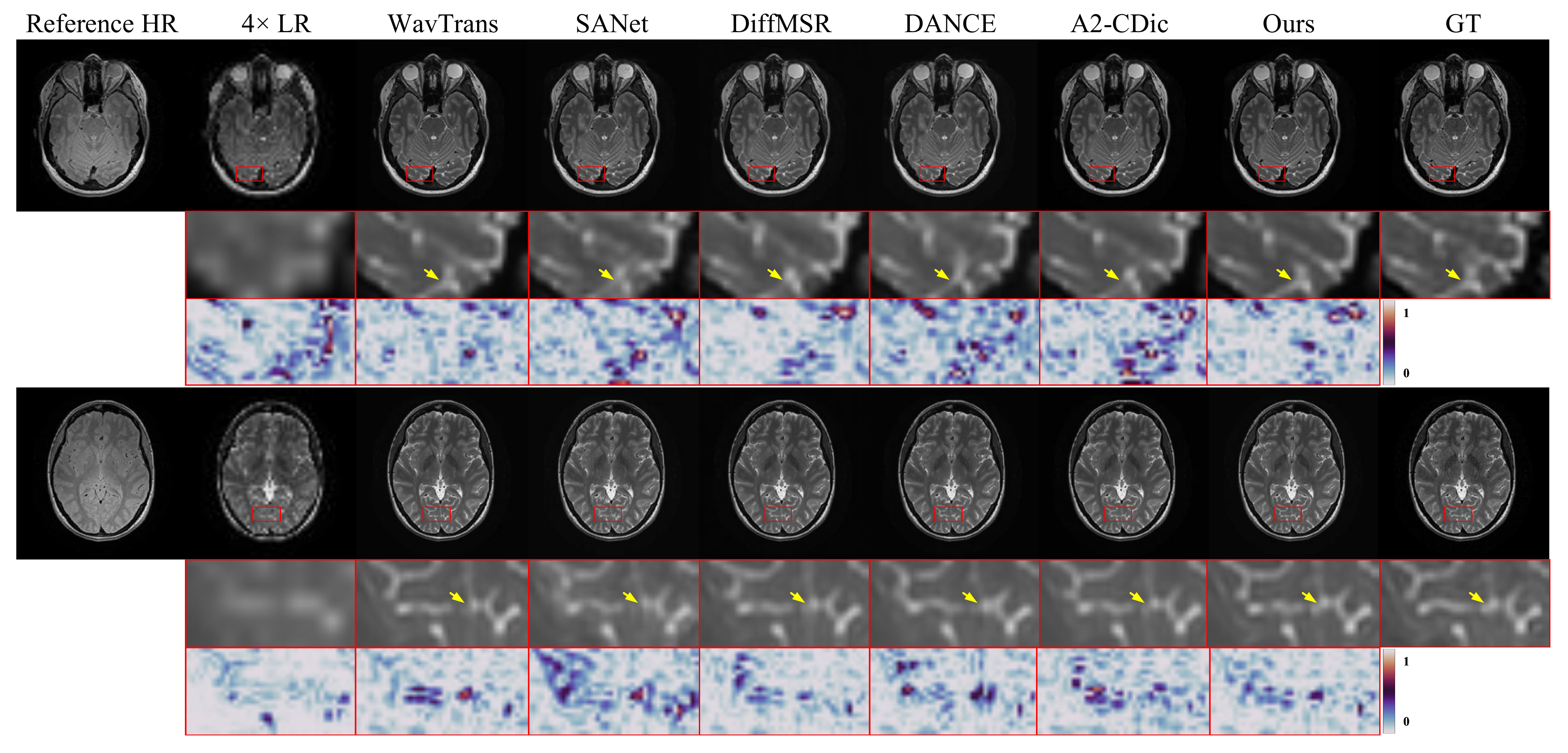}}
	\caption{Qualitative comparison of various methods on the IXI dataset with $4\times$ SR. The yellow arrows indicate areas with significant differences, which are enlarged and shown with residual plots compared to GT.}
	\label{fig_Results_IXI}
\end{figure*}

\section{Results and Analysis}
\subsection{Comparison with Existing Multi-Contrast MRI SR Methods}
\subsubsection{Results on BraTS2018 Dataset}
Table \ref{tab_Results_BraTS} shows a comparison of our method with other methods in terms of super-resolution results on the BraTS2018 dataset. As can be seen, our method achieved the best results in all evaluation metrics for the $2\times$ and $4\times$ super-resolution tasks. Specifically, our method achieves PSNR values of 40.7047 dB and 36.0017 dB, indicating the smallest difference between our results and the ground truth (GT), effectively guiding the restoration of target contrast. The SSIM values reach 0.9885 and 0.9716, respectively, indicating that our method can fully utilize the HR structural information of the reference image.

Figure \ref{fig_Results_BraTS} shows the qualitative comparison results on the BraTS2018 dataset with $4\times$ SR task. As shown in Figure \ref{fig_Results_BraTS}, SANet and DANCE produce smooth results and lose critical texture details. While WavTrans and A2-CDic methods recover structural information, they differ from the GT, resulting in artifacts. In the contrast, Our method can recover complete detail features from LR MRI without causing image distortion or artifacts.

\subsubsection{Results on IXI Dataset}
Table \ref{tab_Results_IXI} presents the quantitative comparison results of super-resolution performance on the IXI dataset, where our approach achieves the highest PSNR values of 43.2223 dB ($2\times$) and 38.5852 dB ($4\times$), significantly outperforming competing methods and demonstrating exceptional fidelity in reconstructed images. Furthermore, our method attains the best structural preservation as evidenced by the top SSIM scores of 0.9876 ($2\times$) and 0.9735 ($4\times$), indicating its outstanding capability in maintaining fine structural details that are crucial for clinical applications. 

Figure \ref{fig_Results_IXI} shows the qualitative comparison results on the IXI dataset with $4\times$ SR task. It can be seen that SANet produce smoother results and lose texture detail information. DiffMSR and DANCE methods generate additional artifacts, leading to incorrect information. Compared to them, our method not only has the smallest difference from GT but also preserves critical texture information.

\begin{table*}[!t]
\centering
\begin{tabular}{lcccc}
\toprule
\multirow{2}[4]{*}{Methods} & \multicolumn{2}{c}{IXI 2$\times$} & \multicolumn{2}{c}{IXI 4$\times$} \\
\cmidrule{2-5}      & PSNR$\uparrow$  & SSIM$\uparrow$  & PSNR$\uparrow$  & SSIM$\uparrow$ \\
\midrule
WavTrans & \underline{42.8824$\pm$2.52} & 0.9852$\pm$0.01 & \underline{38.5073$\pm$2.22} & 0.9711$\pm$0.01 \\
SANet & 41.3978$\pm$2.17 & 0.9825$\pm$0.01 & 35.5812$\pm$2.14 & 0.9417$\pm$0.04 \\
DiffMSR & \multicolumn{1}{c}{/}      & \multicolumn{1}{c}{/}      & 37.4791$\pm$2.33 & 0.9623$\pm$0.03 \\
DANCE & 33.5272$\pm$3.47 & 0.8485$\pm$0.06 & 34.5047$\pm$2.47 & 0.8969$\pm$0.05 \\
A2-CDic & 41.5939$\pm$2.02 & \underline{0.9874$\pm$0.01} & 37.9055$\pm$2.04 & \underline{0.9726$\pm$0.01} \\
\rowcolor[rgb]{ .906,  .902,  .902} CD-DPE & \textbf{43.2223$\pm$2.47} & \textbf{0.9876$\pm$0.01} & \textbf{38.5852$\pm$2.16} & \textbf{0.9735$\pm$0.01} \\
\bottomrule
\end{tabular}%

\caption{Quantitative comparison results of multi-contrast MRI super-resolution on IXI dataset.  \textbf{Bold} indicates the optimal value, while \underline{underline} indicates the second-best value.}
\label{tab_Results_IXI}%
\end{table*}%

\begin{figure}[!t]	\centering{\includegraphics[width=\linewidth]{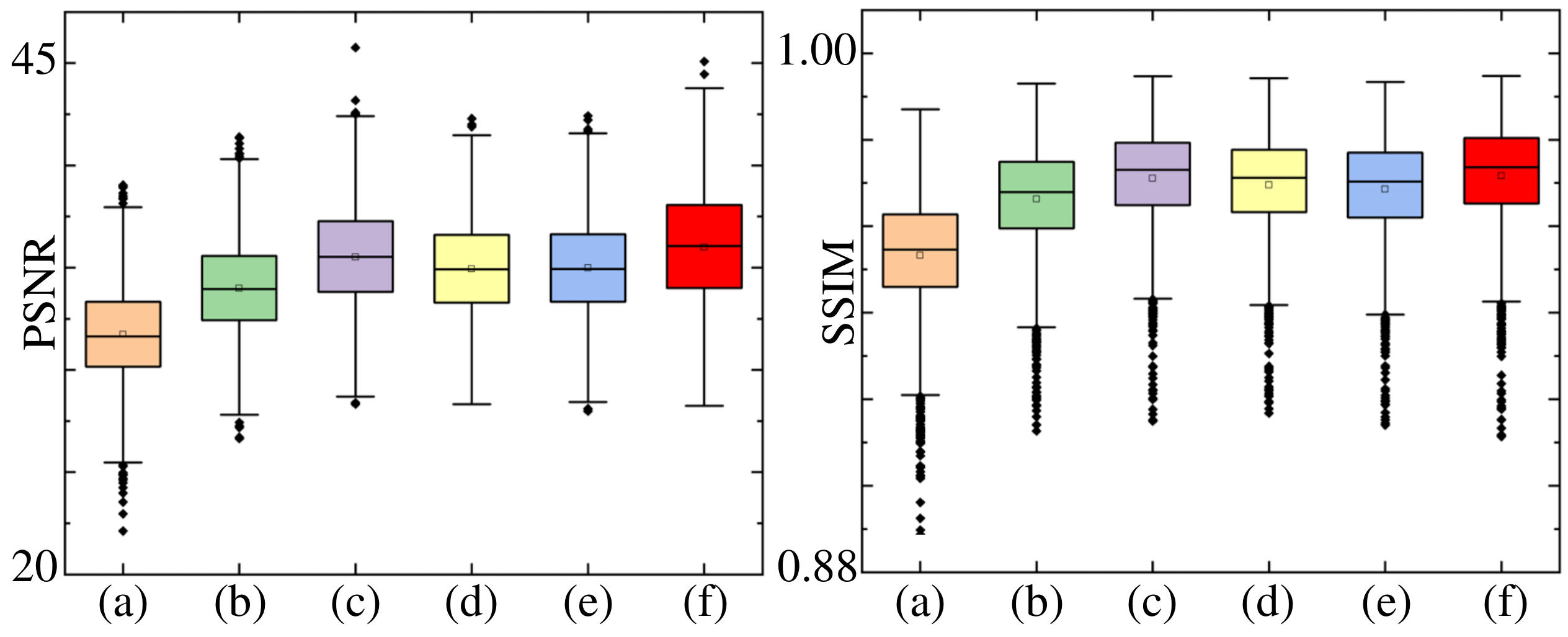}} 
	\caption{Box plots of quantitative results of ablation experiments on the BraTS2018 dataset with $4\times$ SR, where (a)w/o CD-FDM, (b)w/o DP-FFEM, (c)w/o Dual-Prompt, (d)w/o $\mathcal{L}_{mi}$, (e)w/o $\mathcal{L}_{fc}$ and (f)Ours.}
	\label{fig_Results_ablation}
\end{figure}
\begin{figure}[!t]
\centerline{\includegraphics[width=\linewidth]{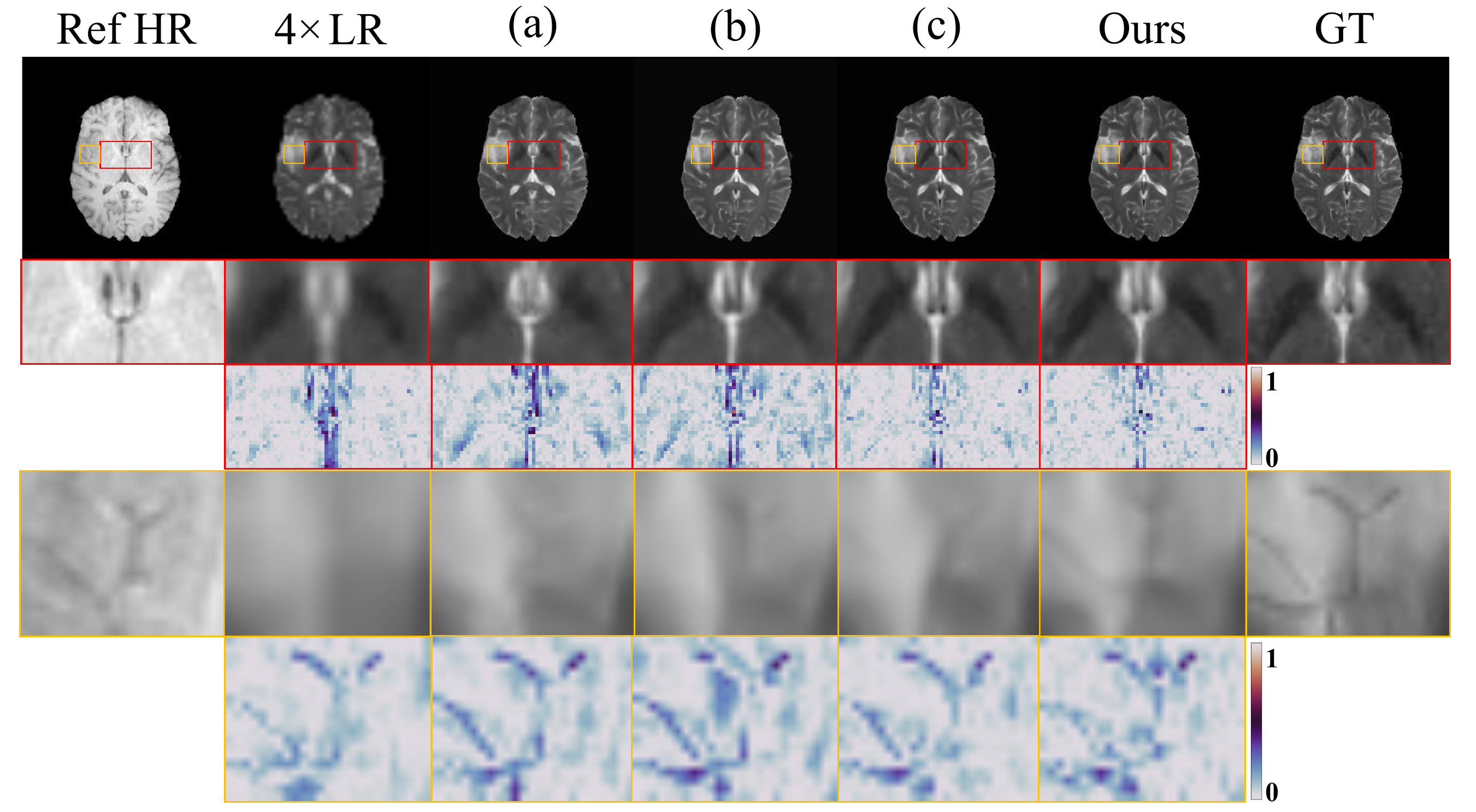}}
	\caption{Qualitative comparison of module ablation experiments on BraTS2018 dataset with $4\times$ SR, where (a)w/o CD-FDM, (b)w/o DP-FFEM, (c)w/o Dual-Prompt.}
	\label{fig_ab_results}
\end{figure}

\begin{figure}[!t]	\centerline{\includegraphics[width=\linewidth]{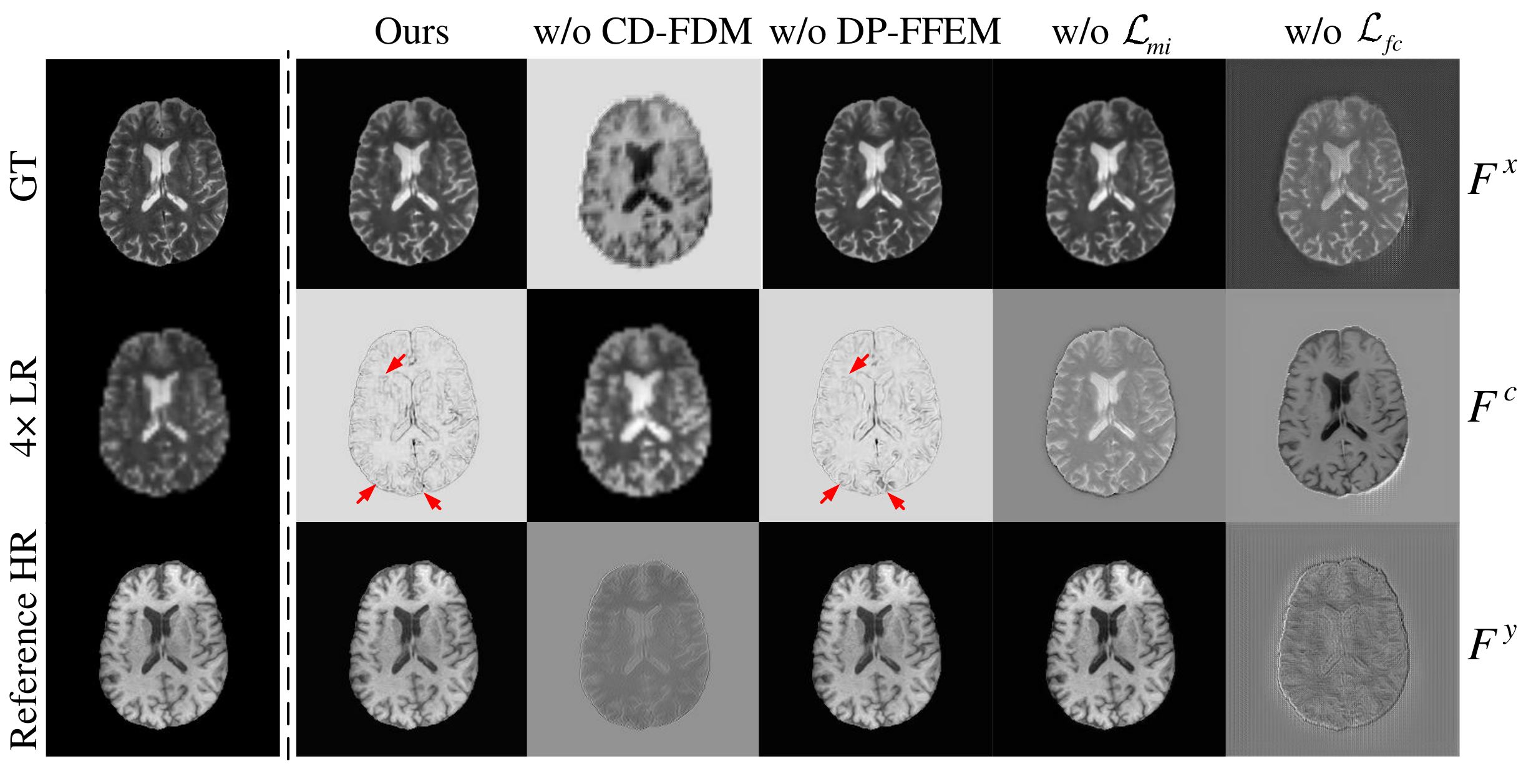}}
	\caption{Visualization of unique and common features in different ablation experiment settings on BraTS2018 dataset with $4\times$ SR.}
	\label{fig_ab_uc_feature}
\end{figure}

\subsection{Ablation Study and Generalizability Analysis}

\subsubsection {Quantitative and Qualitative Ablation Results}
To comprehensively evaluate the contribution of each key component in our CD-DPE framework, we performed systematic ablation studies focusing on CD-FDM, DP-FFEM, dual prompts, and loss functions using BraTS2018 dataset for $4\times$ super-resolution. Quantitative results in Figure \ref{fig_Results_ablation} reveal that: removing CD-FDM (replaced by CNN-based decoupling) caused significant performance drops (13.48\% in PSNR and 1.92\% in SSIM), demonstrating its crucial role in feature extraction; eliminating DP-FFEM (substituted with CNN reconstruction) reduced PSNR by 5.93\% and SSIM by 0.55\%, confirming its effectiveness in frequency-aware enhancement; when removing only dual prompts while retaining DP-FFEM, performance metrics remained inferior to the full model, highlighting the complementary value of prompt mechanisms despite DP-FFEM's standalone effectiveness.

The qualitative ablation results depicted in Figure \ref{fig_ab_results} demonstrate several key observations. The LR image exhibits significant detail loss, while such critical information remains preserved in the reference image. Upon removal of the CD-FDM module, the model fails to effectively extract and utilize reference image features, consequently resulting in unsatisfactory texture detail reconstruction. Elimination of the DP-FFEM module leads to the generation of some details; however, these details show noticeable inconsistency with the GT, accompanied by undesirable artifacts. Interestingly, when dual prompts are excluded from the framework, there is partial mitigation of detail loss, yet the resultant images still suffer from noticeable blurriness. Importantly, our proposed method demonstrates superior performance by maintaining sharp and accurate detail information while effectively avoiding distortions or artifacts in the reconstructed images.

Regarding the impact of loss function components, as demonstrated in Figure \ref{fig_Results_ablation}, the exclusion of $\mathcal{L}_{mi}$ led to performance degradation of 3.05\% in PSNR and 0.22\% in SSIM, while removing $\mathcal{L}_{fc}$ caused reductions of 2.90\% in PSNR and 0.31\% in SSIM. These results confirm that both components play crucial roles in effectively decoupling features and guiding the reconstruction process.

\subsubsection {Effects of Different Components on Unique and Common Features}
Figure \ref{fig_ab_uc_feature} further illustrates how different components affect feature decoupling. When CD-FDM is removed (w/o CD-FDM), the model fails to properly extract the HR reference image's unique features, introducing significant artifacts instead. Moreover, the decomposition of LR target image features becomes flawed, the supposedly unique and common features degenerate into mere intensity inversions that fail to capture the actual structural similarities between reference and target images. This breakdown in feature decomposition highlights the critical role of CD-FDM in maintaining proper feature separation throughout the reconstruction process. 
The loss $\mathcal{L}_{mi}$ enforces disentanglement between shared and unique representations through mutual information minimization. Figure \ref{fig_ab_uc_feature} shows that removing $\mathcal{L}_{mi}$ (w/o $\mathcal{L}_{mi}$) causes feature entanglement, leading to 1.1 dB PSNR drop in reconstruction quality, as validated in Figure \ref{fig_ab_results}. Without using consistency loss $\mathcal{L}_{fc}$ constraint, the model fails to properly disentangle both unique and common features from input images, the subsequent combination of these features cannot accurately reconstruct the original images (Figure 6).

When comparing feature extraction between the model w/o DP-FFEM and our approach, we observed that both successfully separate unique and common features: unique features retain modality-specific contrast, whereas common features capture essential texture details independent of intensity distribution. The key advantage of DP-FFEM lies in its refinement of common features (as indicated by red arrows in Figure \ref{fig_ab_uc_feature}), facilitating robust knowledge transfer from the HR reference to the target image. Crucially, DP-FFEM minimizes reconstruction errors caused by reference-target misalignment. As a result, our method preserves structural details in tumor regions more faithfully, as demonstrated in Figure \ref{fig_ab_results} (comparing w/o DP-FFEM and ours).

\subsubsection {Generalizability on Unseen Dataset}

\begin{table}[!t]
\centering
\begin{tabular}{lcc}
\toprule
\multirow{2}[4]{*}{Methods} & \multicolumn{2}{c}{Generalizability on FastMRI 4$\times$} \\
\cmidrule{2-3}      & PSNR$\uparrow$  & SSIM$\uparrow$ \\
\midrule
WavTrans & \underline{28.0670$\pm$1.83} & 0.7428$\pm$0.04 \\
SANet & 23.0433$\pm$2.70 & 0.5918$\pm$0.07 \\
DiffMSR & 27.3881$\pm$2.58 & 0.7327$\pm$0.08 \\
DANCE & 25.3892$\pm$1.94 & 0.7207$\pm$0.06 \\
A2-CDic & 25.2140$\pm$1.96 & \underline{0.7517$\pm$0.05} \\
\rowcolor[rgb]{ .906,  .902,  .902} CD-DPE & \textbf{29.4134$\pm$2.01} & \textbf{0.8387$\pm$0.04} \\
\bottomrule
\end{tabular}%
\caption{Quantitative results of generalization analysis. All methods were trained on the IXI 4$\times$ dataset and tested on FastMRI Knee 4$\times$ dataset. \textbf{Bold} indicates the optimal value, while \underline{underline} indicates the second-best value.}
\label{tab_Results_Generalizability}%
\end{table}%

\begin{figure}[!t]
\centerline{\includegraphics[width=\linewidth]{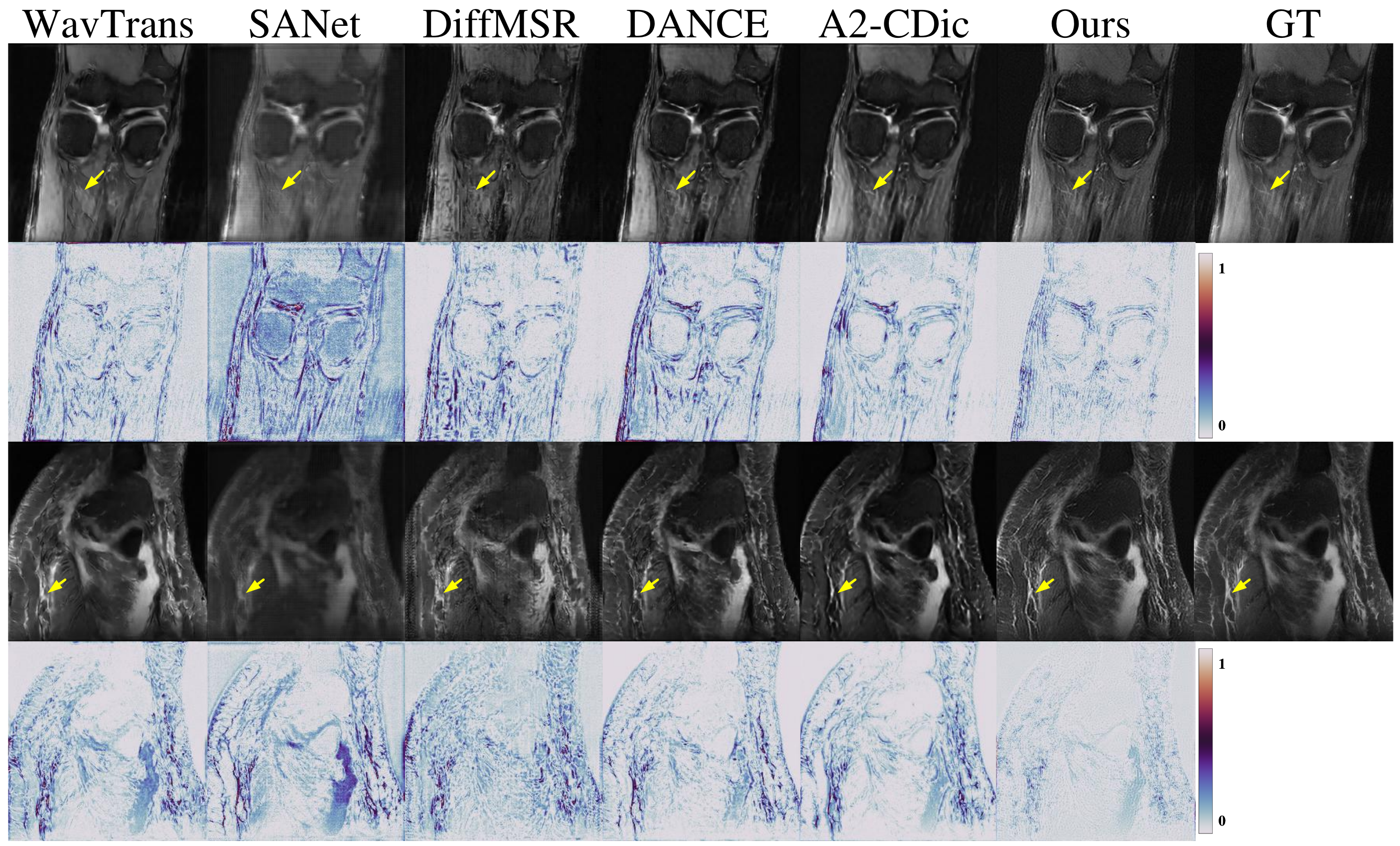}}
	\caption{Qualitative results of generalized analysis. Residual plots relative to GT are displayed.}
	\label{fig_generalization_results}
\end{figure}

To evaluate the adaptability and generalization capability of our proposed method, we conducted direct testing on previously unseen datasets for super-resolution reconstruction. Specifically, a model trained on the IXI 4$\times$ dataset was directly applied to the FastMRI Knee 4$\times$ dataset \cite{knoll2020fastmri}. In the FastMRI dataset, HR PD images serve as reference inputs for reconstructing LR PD-FS target images. A total of 580 test image pairs with a resolution of 256$\times$256 were selected, where the LR inputs were generated via k-space downsampling \cite{lyu2020mri}.

Table \ref{tab_Results_Generalizability} and Figure \ref{fig_generalization_results} report the quantitative and qualitative evaluation results, respectively. As shown, our method exhibits superior generalization performance and robustness across diverse tissue structures and contrast variations in MRI scans, without any additional training. In particular, it achieves improvements of 4.8\% in PSNR and 11.6\% in SSIM compared with the next-best model, demonstrating its strong generalization capability. Furthermore, as illustrated in the residual maps in Figure \ref{fig_generalization_results}, our method yields the smallest deviations from the ground truth while preserving fine texture details. These results indicate that CD-DPE effectively facilitates reference feature fusion and target image reconstruction through the use of dual prompt vectors.

\section{Conclusion}
\label{sec_Conclusion}
In this work, we present CD-DPE model for multi-contrast MRI super-resolution. To tackle the challenges of redundant information and ineffective feature fusion, our approach introduces two key innovations: (1) an iterative CD-FDM to decompose multi-contrast features into cross-contrast and intra-contrast components, eliminating interference while preserving structural details; and (2) a DP-FFEM that adaptively integrates complementary information through frequency-aware feature selection and dynamic routing-based fusion. Extensive experiments on public datasets demonstrate that CD-DPE significantly enhances reconstruction accuracy, recovering fine anatomical structures with reduced artifacts and superior sharpness compared to existing methods. The ablation studies also validate the effectiveness of the proposed CD-FDM and DP-FFEM. Additionally, CD-DPE demonstrated strong generalization capabilities when validated on unseen datasets

\textbf{Limitations} While CD-DPE demonstrates superior reconstruction accuracy, limitations include: (1) the model remains sensitive to extreme contrast discrepancies between reference and target images, particularly when the contrast mechanisms substantially differ. Future research should explore incorporating MRI physics principles, such as quantitative relaxation mapping or biophysical models, to better bridge such contrast differences; (2) the iterative nature of the feature decoupling process introduces computational overhead. Developing more efficient mechanisms for unique and common feature extraction that eliminate the need for unfolding-based learning represents an important direction for future work.


\section*{Acknowledgements}
This work was supported by the National Natural Science Foundation of China (Grant No.62571150), and the Guizhou Provincial Basic Research Program (QianKeHe ZK [2023] 058).

\bibliography{aaai2026}

\end{document}